# FilMBot: A High-Speed Soft Parallel Robotic Micromanipulator

Jiangkun Yu, Houari Bettahar, Hakan Kandemir, and Quan Zhou

***Abstract*—Soft robotic manipulators are generally slow despite their great adaptability, resilience, and compliance. This limitation also extends to current soft robotic micromanipulators. Here, we introduce FilMBot, a 3-DOF film-based, electromagnetically actuated, soft kinematic robotic micromanipulator achieving speeds up to 2117 °/s and 2456 °/s in α and β angular motions, with corresponding linear velocities of 1.61 m/s and 1.92 m/s using a 4-cm needle end-effector, 0.54 m/s along the Z axis, and 1.57 m/s during Z-axis morph switching. The robot can reach ~1.50 m/s in path-following tasks, with an operational bandwidth below ~30 Hz, and remains responsive at 50 Hz. It demonstrates high precision (~6.3 µm, or ~0.05% of its workspace) in path-following tasks, with precision remaining largely stable across frequencies. The novel combination of the low-stiffness soft kinematic film structure and strong electromagnetic actuation in FilMBot opens new avenues for soft robotics. Furthermore, its simple construction and inexpensive, readily accessible components could broaden the application of micromanipulators beyond current academic and professional users.**

## I. INTRODUCTION

ROBOTIC micromanipulation is a crucial technology for a wide range of applications that impact our lives, from the fabrication of precision sensors to biomedical interventions. For instance, robotic micromanipulation is a key tool in assembling complex micro-optoelectromechanical systems (MOEMS), such as microspectrometers [1], waveguides and resonators [2]. Micromanipulators are also an indispensable means for cell manipulation, including the handling of individual biological cells [3] and intracytoplasmic sperm injection [4]. As robotic micromanipulation continues to evolve, it pushes the boundaries in both scientific research and biomedical and industrial applications.

While traditional rigid micromanipulators have played a vital role in precision manipulation tasks, they often struggle with compliance [5], [6]. In response, soft robotic micromanipulators have been proposed, offering great flexibility and resilience [7], [8]. However, these soft systems lack speed and precision for dynamic applications. Many soft robotic micromanipulators have only reported their quasi-static performances [9], [10]. For a few that reported their dynamic performances, the speeds are very limited. For example, the 3-degree-of-freedom (DOF) robotic micromanipulator actuated by piezohydraulic actuators and bellows [11], probably the first soft parallel micromanipulator, has achieved a velocity of 0.8 mm/s, with an exceptional precision of ±0.5 µm, about 0.2 % of the size of its workspace. The 3-DOF soft parallel robot employing dielectric elastomer actuators can operate at a maximum speed of 1.5 mm/s and has a precision of 13 µm, around 0.4 % of its workspace length [12]. Another 3-DOF manipulator made of silicone rubber tubing, with its great dimensions of 153 mm, has reached a higher speed of about 40 mm/s under high-pressure air drive, at a precision of 0.62 to 0.94 cm, which is approximately 2.6 % to 3.9 % of the size of its workspace [13]. Micromanipulators with shape memory alloy (SMA) as actuators usually have lower speeds due to their large response delays [14], although the precision could be similar [15].

In contrast, state-of-the-art rigid micromanipulators can achieve speeds orders of magnitude higher, such as the piezoelectric-driven milliDelta [16] and MiGriBot [17]. The milliDelta, for instance, reached a speed of 450 mm/s, at a precision of ~5 µm, approximately 0.16 % of the workspace length. The MiGriBot can maintain a speed of 60 mm/s, at a better precision of 1 µm, or 0.1 % of its workspace length. In addition, a fiberglass-based continuum micromanipulator actuated by electric motors achieves a maximum speed of 65 mm/s and a precision of about 0.31 mm, or 1.9 % of the length of its workspace [18].

In this paper, we introduce FilMBot, a 3 DOF high-speed, high-bandwidth, and high-precision soft kinematic robotic micromanipulator employing a novel combination of film-based parallel kinematics and contactless electromagnetic actuation. The centimeter-scale FilMBot achieves a high speed of up to 2117 °/s, 2456 °/s and 0.54 m/s in α, β, and Z axis, respectively, and 1.57 m/s in Z-axis morph switching. With a 4-cm needle end-effector, it can reach corresponding linear velocities of 1.61 m/s and 1.92 m/s along the X and Y axes, and an average velocity of ~1.50 m/s in path-following tasks. It demonstrates a high precision of ~6.3 µm for path following, which corresponds to about 0.05 % of its workspace size. Furthermore, it has an operational bandwidth below ~30 Hz and remains responsive at 50 Hz.

This work was supported by the Academy of Finland grant 355769 and Research Council of Finland grant 362715. (Corresponding author: *Quan Zhou*.)

Jiangkun Yu, Houari Bettahar, and Quan Zhou are with the Department of Electrical Engineering and Automation, Aalto University, 02150 Espoo, Finland (e-mail: jiangkun.yu@aalto.fi; houari.bettahar@aalto.fi; quan.zhou@aalto.fi).

Hakan Kandemir is with the Microelectronics and Quantum Technology, VTT Technical Research Centre of Finland, 02150 Espoo, Finland (e-mail: hakan.kandemir@vtt.fi).

This article has supplementary material provided by the authors.

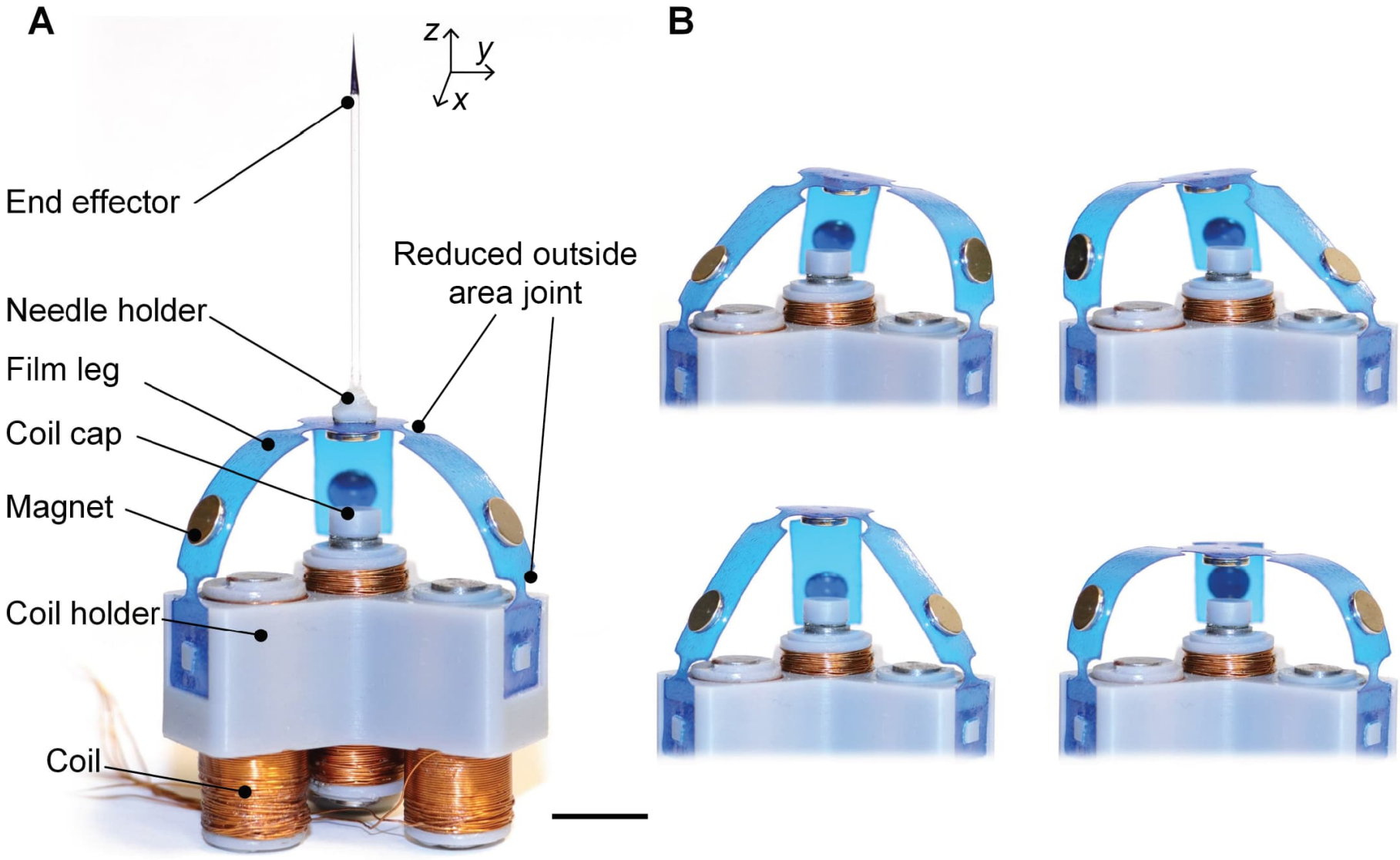

**Fig. 1.** FilMBot: a film-based soft parallel robotic micromanipulator. (A) A prototype of the FilMBot with components labeled, scale bar 10 mm. The FilMBot consists of four magnet coil pairs, a film-based soft kinematic structure, and 3D-printed structural parts (a coil holder, a needle holder, and a coil cap). A needle end-effector is attached to observe both the translational and rotational motion of the FilMBot. (B) Perspective views of the FilMBot near the right, left, top, and bottom boundaries of its workspace.

The rest of this article is organized as follows. The design and kinematic modeling are shown in Section II. The quasi-static and dynamic performance of the FilMBot is presented in Section III. Section IV demonstrates the capability of the FilMBot in puncturing. Finally, Section V presents the conclusion and discussion.

## II. Design and Modeling

### A. Working Principle

The design of the FilMBot is inspired by the tripod-like manipulator [19] [20] as well as previous work [11], which achieved kinematic softness from rigid components. To achieve high speed, we adopted film for the soft kinematic structure, actuated with strong magnetic coupling between steel-core solenoid coils and rare-earth permanent magnets, illustrated in Fig. 1. The low stiffness of the soft kinematic structure allows rapid deformation, while the combination of electromagnetic coils and permanent magnets ensures fast responses and strong forces. The entire kinematic structure is fabricated from a single piece of polypropylene material. To reduce longitudinal bending and transverse bending stiffness of the legs, reduced outside area joints [21], or neck joints, were introduced at their upper and lower ends. Each leg is actuated by a rigid permanent magnet mounted on its outer side. Rare-earth permanent magnets were employed for their high magnetic response and capability to both be pushed and pulled magnetically. Three steel-core solenoid coils were installed under each leg to actuate the respective legs. To improve actuation performance in the Z direction, an additional permanent magnet was mounted below the top platform where the three legs converge, and actuated by the fourth redundant central coil placed underneath. A plastic cap atop the core of the central coil prevents the top magnet from falling further and locking there (see Appendix A). The kinematic structure of the proposed FilMBot is 5 cm in height and 2.2 cm in circumradius. Additionally, a needle holder was affixed onto the top platform, where a 4-cm-long needle was installed to facilitate the observation of the motion of the FilMBot and serve as the end-effector.

By varying the magnetic field generated by the solenoid coils, the magnets deform the soft legs, causing the top platform to move and rotate, achieving various spatial motions, as illustrated in Fig. 1B. The motion of the FilMBot in response to varying current inputs is shown in Appendix B and Supplementary Movie 1.

### B. Fabrication

The FilMBot was constructed using accessible materials and easily fabricated components. The film legs and top platform were laser-cut from a single piece of recyclable polypropylene (PP) film (Office DEPOT, cut flush folders A4), with Young's modulus 943 MPa ± 15 MPa (n = 6). The legs were mounted with round NdFeB permanent magnets (Supermagnet, S-06-0.75-STIC) of size ϕ6 mm × 0.75 mm. The coils were hand-wound with 3D-printed shoulders at both ends of the ϕ6-mm steel core (Warma Steel, S355J2 rod steel) to ensure uniform winding. The end-effector needle was made from a ϕ1-mm glass capillary with the tip heated and hot-drawn to a thinner profile. The coil holder, coil cap, and needle holder were designed using SolidWorks (Version 2022) and 3D printed with a Prusa SL1S printer using Liqcreate Strong X resin.

To assemble the base part of the FilMBot, three surrounding coils were fitted into the coil holder, with gaps filled with copper foil to ensure a snug fit. The coils were leveled and aligned using a spirit level. The central coil, positioned higher than the surrounding coils, was installed using the same method. The coil cap was then secured on top of the central coil core with double-sided tape.

To precisely mount the magnets, a custom-built 3 DOF robotic arm was employed, along with a 3 DOF platform. An iron cylinder head was attached to the 3 DOF robotic arm to transport the permanent magnet onto the desired locations on the film legs and the top platform fixed on the 3 DOF platform, with assistance from the top and side view cameras.

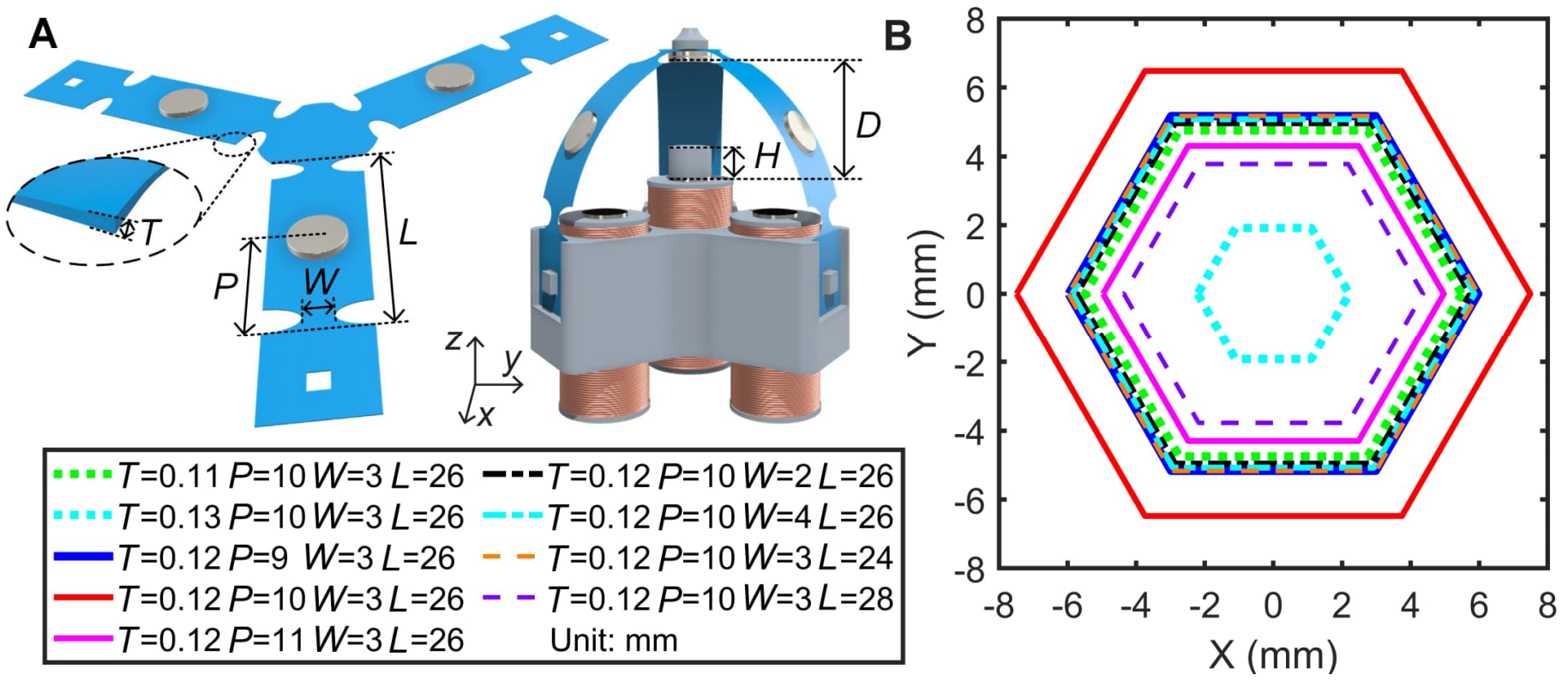

**Fig. 2.** Design parameters and workspace comparison. (A) Visual annotation of design parameters, including film thickness *T*, magnets position *P*, joint width *W*, and leg length *L*. (B) XY plane projections of the workspace corresponding to different parameters.

The three film legs were mounted on the coil holder using precut holes that anchor onto protruded bumps and secured using double-sided tape. The glass needle was secured to a 3D-printed needle holder using superglue and affixed to the center of the top platform with double-sided tape. The fabrication procedure mentioned above is demonstrated in Supplementary Movie 2.

### *C. Design Parameters*

To achieve high speed, we employed polypropylene film for the soft structure and strong magnetic forces between the steel-core solenoid coils and the permanent magnets for actuation. The solenoid coils, with a 6-mm core diameter matching the size of the permanent magnets, ensured a stable geometrical configuration when the magnet was drawn in and made contact with the top of the steel core. The coil diameter was maximized within the constraints of the dimension of the micromanipulator to enhance the magnetic field strength. Other key parameters are shown in Fig. 2.

For high-speed response, the film should have a low stiffness, in the meanwhile, the film should also be strong enough to support components such as permanent magnets and film itself. For that, we employed a simplified bridge-like film structure loaded with a dummy load of the weight of the magnet to determine the film thickness (see Fig. 3). The simulation was performed using COMSOL Multiphysics (Version 6.1), and the relationship between the film thickness and the film displacement caused by gravity was obtained at different film thicknesses. Fig. 3B shows that the structure with film thicknesses of 100 μm and 200 μm deforms about 1 % and 0.1 %, respectively. On the other hand, the maximum achievable force through electromagnetic actuation in the work distance is approximately 185 mN (see Fig. 4C), about 100 times the gravity of a magnet (0.1612 g). Consequently, to keep the gravity-induced displacement relatively small, and ensure magnetic force can induce significant displacement, we favor a candidate film thickness range of 100 μm to 200 μm, preferably at the low end for its reduced stiffness.

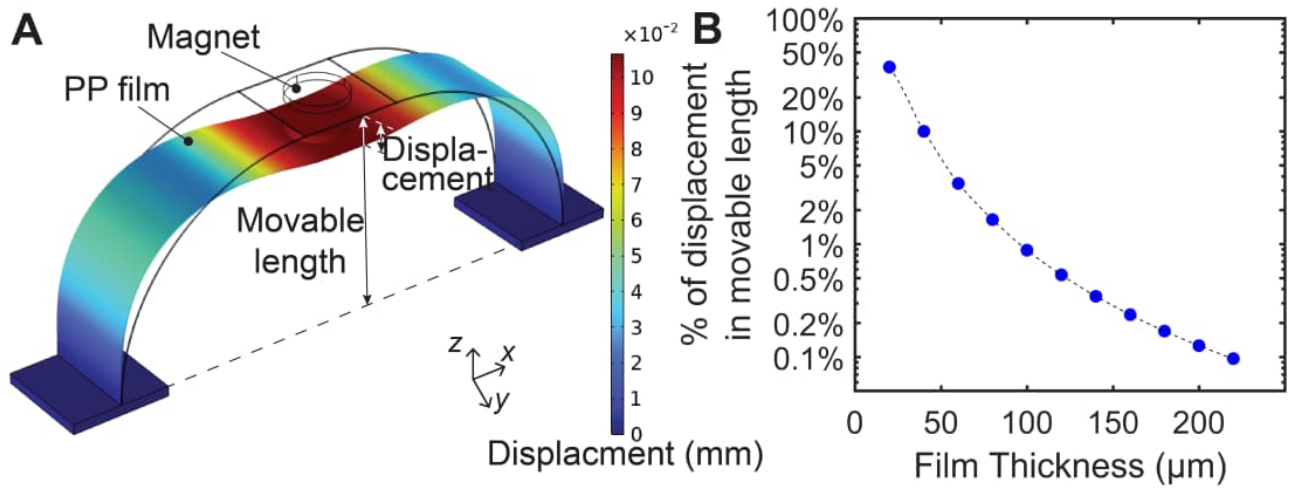

**Fig. 3.** Simulation study of the effect of film thickness in a simplified bridge-like film structure. (A) The structure with a magnet load. (B) Percentage of displacement at different film thickness.

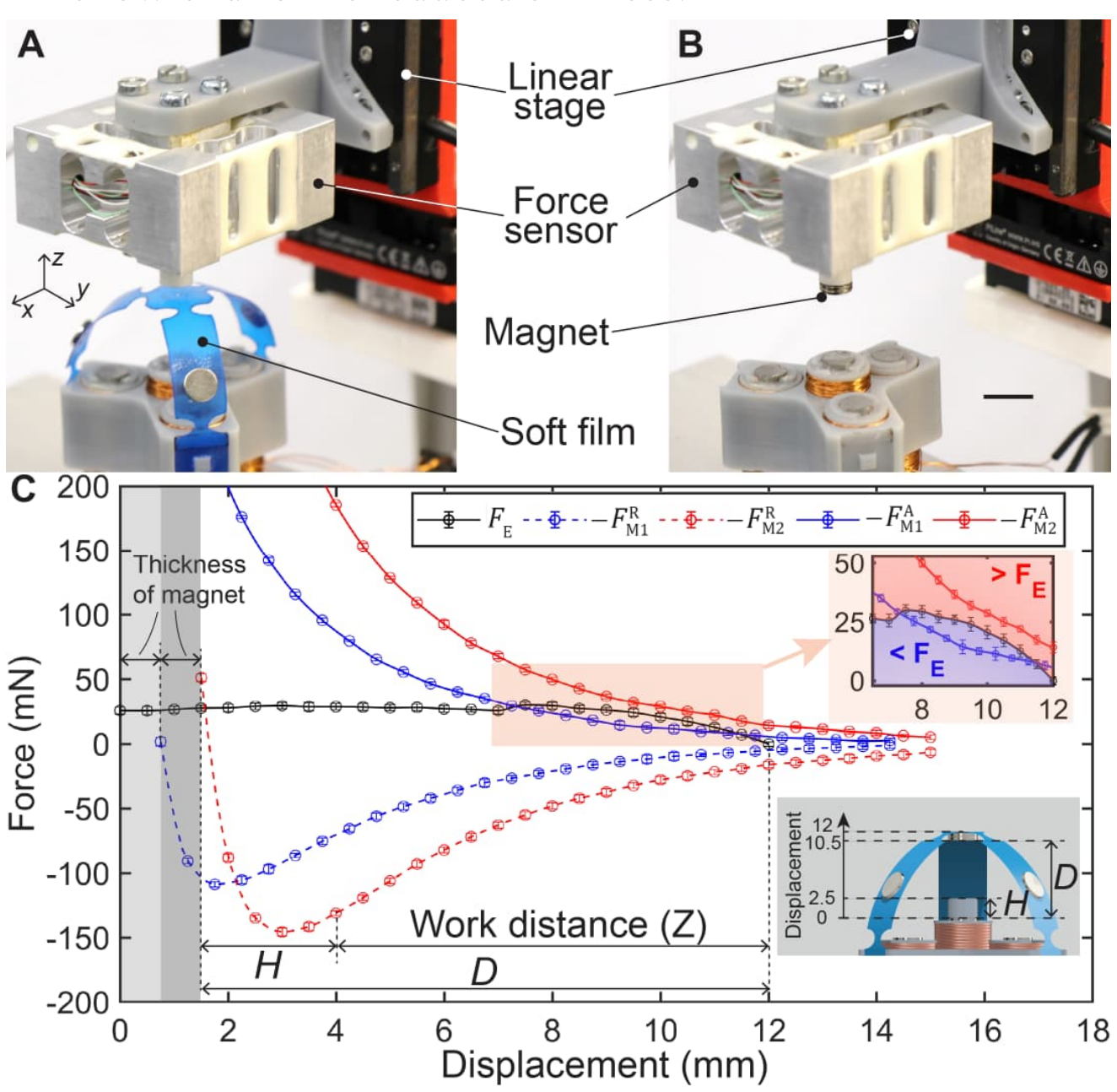

**Fig. 4.** Experimental analysis of elastic and magnetic forces in the soft kinematic structure. (A) Setup for measuring the elastic force, with the top magnets and coil cap removed. (B) Setup for measuring magnetic force, with the film kinematic structure removed, magnets attached to the sensor, and maximum current applied to the central coil. The scale bar is 10 mm. (C) Comparison of the measured elastic force $F_{\mathrm{E}}$, the magnetic attractive force $F_{\mathrm{M}}^{\mathrm{A}}$ and the magnetic repulsive force $F_{\mathrm{M}}^{\mathrm{R}}$, with subscripts 1 and 2 indicating the number of top magnets. The displacement axis indicates the distance between the central coil top and the top platform. The gray shaded area represents the thickness of one or two magnets, which reduces the minimum work distance in the Z axis.

Within this range, we experimentally determined the film thickness *T* and other parameters, including magnet position *P*, joint width *W*, and leg length *L*, as labeled in Fig. 2A, using workspace as a performance criterion. The parameters were determined iteratively (see Algorithm 1 and Appendix C), and the corresponding workspaces for different parameters are shown in Fig. 2B. The resulting final parameters were $T = 0.12$ mm, $P = 10$ mm, $W = 3$ mm, and $L = 26$ mm.

Additionally, the separation distance between the central coil and the top magnets *D*, the height of the central coil cap *H*, and the number of stacked top magnets *M* significantly influence the FilMBot performance along the Z axis. We experimentally measured the elastic force from the deformation of the soft kinematic structure and the magnetic force between the central coil and the top magnets, as shown in Fig. 4, to evaluate their effects on the Z-axis performance.

To ensure that the magnet can be drawn in by the central coil, the magnetic attractive force should be greater than the elastic force. As shown in Fig. 4C, the magnetic attractive force of a single magnet $F_{\mathrm{M1}}^{\mathrm{A}}$ (blue solid line) is smaller than the elastic force $F_{\mathrm{E}}$ (black solid line) at some displacement. Adjusting the position of the central coil vertically could improve this but would reduce the work distance on the Z axis. Therefore, a second magnet was added to ensure the attractive force ($F_{\mathrm{M2}}^{\mathrm{A}}$, red solid line) consistently greater than $F_{\mathrm{E}}$.

However, when the top magnets are drawn in and contact the coil core, they may lock there due to the magnetic repulsive force $F_{\mathrm{M2}}^{\mathrm{R}}$ being stronger than $F_{\mathrm{E}}$ at close proximity (detailed in Appendix A). To ensure that the top magnets can be pushed away, a 3D-printed coil cap was mounted on the center magnet to maintain a minimum 0.5-mm gap between the top magnets and the center coil. However, with $H = 0.5$ mm, the top magnets will tilt and flip to the side of the core when the magnet pushes them away. Therefore, we selected *H* to be 2.5 mm, making the system more robust and stable in practice. The final parameters were $D = 10.5$ mm, $H = 2.5$ mm, and $M = 2$.

*D. Kinematic Model*

To estimate the workspace of the FilMBot, we developed a kinematic model using experimental input-output data. This empirical approach implicitly captures system-level effects, including magnetic field nonlinearities and cross-magnetization between coils and magnets. Let $\mathbf{i} = [i_1 \quad i_2 \quad i_3 \quad i_4]^T$ be the input currents vector for the four coils, where $^T$ denotes the transpose, and $\mathbf{p}_{\mathrm{p}}$ and $\mathbf{p}_{\mathrm{n}}$ are Cartesian coordinates vectors of the output positions for the top platform and the needle tip, respectively. The relationship between the system inputs and outputs is given by:

$$\mathbf{X} = f(\mathbf{i}) \tag{1}$$

where $\mathbf{X} = [\mathbf{p}_{\mathrm{p}} \quad \mathbf{p}_{\mathrm{n}}]^T$. The function $f$ was derived from experimental data. Due to the nonlinear magnetic field and kinematics of the soft structure, both first-order and second-order current values were used as the explaining variable for the regression of the direct kinematic model:

$$\mathbf{X} = \mathbf{A}\mathbf{V} + \mathbf{B} \tag{2}$$

**Algorithm 1:** Iteratively selecting parameters $T$, $P$, $W$, $L$.

**In:** Initial guess values: $T_{\mathrm{g}} = 110$ μm, $P_{\mathrm{g}} = 10$ mm, $W_{\mathrm{g}} = 3$ mm, $L_{\mathrm{g}} = 26$ mm.
Searching step size: $T_{\mathrm{s}} = 10$ μm, $P_{\mathrm{s}} = 1$ mm, $W_{\mathrm{s}} = 1$ mm, $L_{\mathrm{s}} = 2$ mm.

**Out**: Selected parameters $T_{\mathrm{b}}$, $P_{\mathrm{b}}$, $W_{\mathrm{b}}$, and $L_{\mathrm{b}}$.

1: Select film thickness $T$:
  **for** $T$ in $[T_{\mathrm{g}} - T_{\mathrm{s}}, T_{\mathrm{g}}, T_{\mathrm{g}} + T_{\mathrm{s}}]$ **do**
    Set $P = P_{\mathrm{g}}$, $W = W_{\mathrm{g}}$, $L = L_{\mathrm{g}}$
    Perform experiments and get workspace $S$
  $T_{\mathrm{b}} \leftarrow \arg\max_{T} S(T)$

2: Select magnets position $P$:
  **for** $P$ in $[P_{\mathrm{g}} - P_{\mathrm{s}}, P_{\mathrm{g}}, P_{\mathrm{g}} + P_{\mathrm{s}}]$ **do**
    Set $T = T_{\mathrm{b}}$, $W = W_{\mathrm{g}}$, $L = L_{\mathrm{g}}$
    Perform experiments and get workspace $S$
  $P_b \leftarrow \arg\max_{P} S(P)$

3: **If** $P_{\mathrm{b}} \neq P_{\mathrm{g}}$ **then**
    $P_{\mathrm{g}} = P_{\mathrm{b}}$
    Go to step 1

4: Select joint width $W$:
  **for** $W$ in $[W_{\mathrm{g}} - W_{\mathrm{s}}, W_{\mathrm{g}}, W_{\mathrm{g}} + W_{\mathrm{s}}]$ **do**
    Set $T = T_{\mathrm{b}}$, $P = P_{\mathrm{b}}$, $L = L_{\mathrm{g}}$
    Perform experiments and get workspace $S$
  $W_b \leftarrow \arg\max_{W} S(W)$

5: **If** $W_{\mathrm{b}} \neq W_{\mathrm{g}}$ **then**
    $W_{\mathrm{g}} = W_{\mathrm{b}}$
    Go to step 1

6: Select leg length $L$:
  **for** $L$ in $[L_{\mathrm{g}} - L_{\mathrm{s}}, L_{\mathrm{g}}, L_{\mathrm{g}} + L_{\mathrm{s}}]$ **do**
    Set $T = T_{\mathrm{b}}$, $P = P_{\mathrm{b}}$, $W = W_{\mathrm{b}}$
    Perform experiments and get workspace $S$
  $L_{\mathrm{b}} \leftarrow \arg\max_{L} S(L)$

7: **If** $L_{\mathrm{b}} \neq L_{\mathrm{g}}$ **then**
    $L_{\mathrm{g}} = L_{\mathrm{b}}$
    Go to step 1

8: **return** $T_{\mathrm{b}}$, $P_{\mathrm{b}}$, $W_{\mathrm{b}}$, and $L_{\mathrm{b}}$

where $\mathbf{V} = [\mathbf{i} \quad \mathbf{i}^2]^T$, $\mathbf{i}^2 = [i_1^2 \quad i_2^2 \quad i_3^2 \quad i_4^2]^T$. $\mathbf{A}$ and $\mathbf{B}$ are the coefficient and offset matrices, respectively. The inverse kinematic model can be derived using (2).

The coefficient and offset matrices were identified from independent static samples. For each current vector $\mathbf{i}$ drawn from a randomized set, the robot was starting from zero current, allowed to reach equilibrium, and the positions of two points on the needle were recorded, including the needle tip, which are denoted as $\mathbf{p}_{\mathrm{m}}$ and $\mathbf{p}_{\mathrm{n}}$, respectively, as shown in Fig. 5. The position $\mathbf{p}_{\mathrm{p}}$ was then calculated using the following geometric relation:

$$\mathbf{p}_{\mathrm{p}} = \mathbf{p}_{\mathrm{n}} - l_{\mathrm{n}}\mathbf{u} \tag{3}$$

where $\mathbf{u} = \frac{\mathbf{p}_{\mathrm{n}} - \mathbf{p}_{\mathrm{m}}}{\|\mathbf{p}_{\mathrm{n}} - \mathbf{p}_{\mathrm{m}}\|}$ is the unit vector along the direction of $\mathbf{p}_{\mathrm{m}}$ to $\mathbf{p}_{\mathrm{n}}$, and $l_{\mathrm{n}}$ is the needle length. Then, regression was performed with multiple sets of experimental data of $\mathbf{p}_{\mathrm{p}}$, $\mathbf{p}_{\mathrm{n}}$,

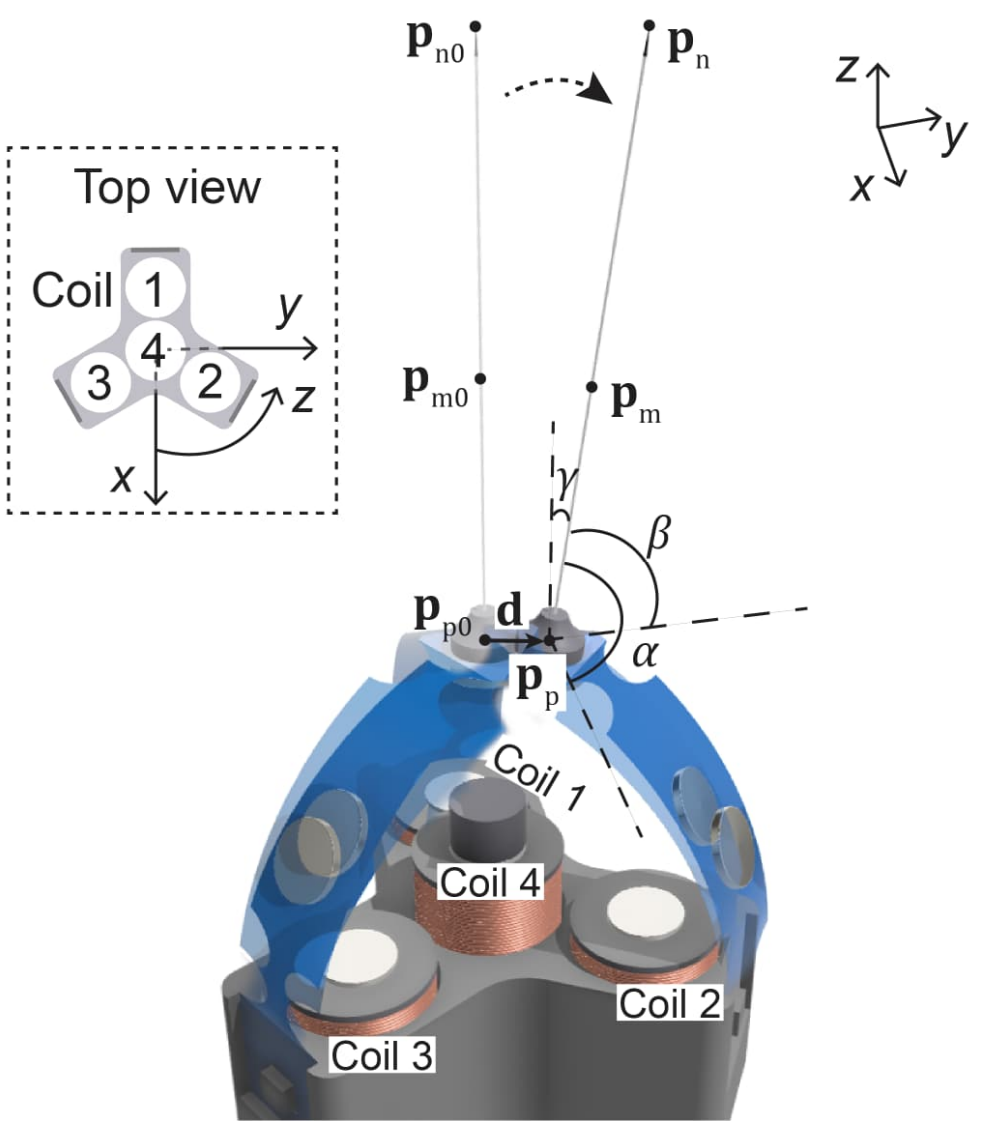

**Fig. 5** Schematic of the coordinates and points during the movement of the FilMBot, as well as the configuration of coils.

and **i** to identify the coefficient matrices **A** and **B** using least squares.

In addition, the motion of the FilMBot includes both translations and rotations. Let the translation **d** and rotation **r** of the top platform be $\mathbf{d} = [d_\mathrm{x} \quad d_\mathrm{y} \quad d_\mathrm{z}]^T$ and $\mathbf{r} = [\alpha \quad \beta \quad \gamma]^T$, where $d_\mathrm{x}$, $d_\mathrm{y}$ and $d_\mathrm{z}$ represents translations along the X, Y, and Z axes, and $\alpha$, $\beta$ and $\gamma$ are the rotation angles of the needle relative to the X, Y, and Z axes, respectively, we have:

$$\mathbf{d} = \mathbf{p}_\mathrm{p} - \mathbf{p}_\mathrm{p0} \tag{4}$$

$$\mathbf{r} = \left[\cos^{-1}(u_\mathrm{x}) \quad \cos^{-1}(u_\mathrm{y}) \quad \cos^{-1}(u_\mathrm{z})\right]^T \tag{5}$$

where $\mathbf{p}_\mathrm{p0}$ is the initial position of the top platform, obtained experimentally when $\mathbf{i} = \mathbf{0}$. $u_\mathrm{x}$, $\mathrm{u}_\mathrm{y}$, and $\mathrm{u}_\mathrm{z}$ are the components of **u** on the X, Y, and Z axes, respectively. While the 3D orientation of the needle is used to calculate $\alpha$, $\beta$, and $\gamma$, only $\alpha$ and $\beta$ (rotation around the X and Y axes) are actively controlled. FilMBot does not actuate or utilize rotation around the Z-axis, and $\gamma$ is included only for completeness of geometric representation and consistency in notation.

## III. Performances

### A. Quasi-Static Performance

To evaluate the quasistatic performance of the FilMBot, a series of experiments were conducted using custom-built electronics. The coils of the FilMBot are connected to a custom-printed circuit board with motor drivers (SparkFun, Dual TB6612FNG), which are controlled by PWM signals from the microcontroller (Arduino, Mega 2560). The current of the coils is measured using shunt resistors for both closed-loop current control at 1 kHz and recording to a computer via USB. Two frame-synchronized high-speed cameras (Phantom, Miro LC310) were set up orthogonal to record experiments. A custom Python script (python 3.6 with opencv-python 4.6) was used to identify the needle position.

The film-based soft kinematics enables significant deformations in the structure of the FilMBot. The magnets can also be drawn in by strong magnetic forces and adhere to the coil core, which will interestingly create alternative morphs of the FilMBot. In its default state, without any magnet adhering to the coil core, the FilMBot maintains its primary form, or main morph, as shown in Fig. 6A. When a magnet adheres to the coil core, the kinematic structure is significantly altered, switching the FilMBot to a side morph, as exhibited in Fig. 6B. In these side morphs, the remaining unattached magnet-coil pairs continue to function, due to the actuation redundancy of the system, resulting in distinct workspaces. Step signals were typically used to switch from the main morph to the side morphs, with each transition completed within 0.05 s (see Appendix B), during which the trajectory was uncontrollable.

To characterize the main morph, we measure the tip displacement of a 4-cm-long end effector by applying 4096 linear combinations of four coil currents, limited in [-0.5 A, 0.4 A] to prevent the magnet from being drawn in. The measured displacement of the end effector and the workspace boundary estimated by the kinematic model from Section II are depicted in Fig. 6C. The end effector can rotate approximately 18.5° and 18.4° relative to the X and Y axes, respectively. The length of the main workspace along the Z axis is approximately 3.51 mm. If the top magnets move lower, they will be drawn in, causing the FilMBot to switch to morph B4. The corresponding length of the workspace reaches maximum 14.97 mm and 12.96 mm in the X and Y directions, respectively, for the 4-cm needle tip. The main workspace is hexagonal in all XY, XZ, and YZ views, and the estimated 3D workspace is illustrated in Fig. 6E.

The quasistatic workspaces of the side morphs of the FilMBot were measured similarly. The overall configuration of the workspaces of the main morph and side morphs is shown in Fig. 6D. Notably, the three surrounding side-morph workspaces are similar, each taking on a wing-like shape and forming angles of approximately 120° relative to each other, mirroring the angles between the film legs of the FilMBot. As one leg is pulled in, the needle tilts and descends, causing the workspaces of side morphs to be lower than the workspace of the main morph. Although the entire workspace is an unconventional shape with three wings, the multimorph characteristic yields a significantly expanded workspace compared to the main workspace alone. The maximum reach extends to 41.47 mm along the X axis, 52.62 mm along the Y axis, and 11.22 mm along the Z axis. Supplementary Movie 3 and Appendix B demonstrate the operation of the FilMBot in side morphs and morph switching.

Additionally, we measured the blocked force with a force sensor (ME-Meßsysteme, K3D40) mounted next to the FilMBot and the maximum current applied to coils. For the Z direction, the measurement region corresponds to the length of the workspace when switching between the B4 morph and the main morph. The maximum blocked force measured was 235 mN ± 12 mN, as depicted in Fig. 7A, while in the main morph, the maximum force was 79 mN ± 9 mN. In addition, the maximum blocked forces along the X and Y axes measured were 10.8 mN ± 0.5 mN and 8.9 mN ± 0.4 mN, respectively, as depicted in Fig. 7B.

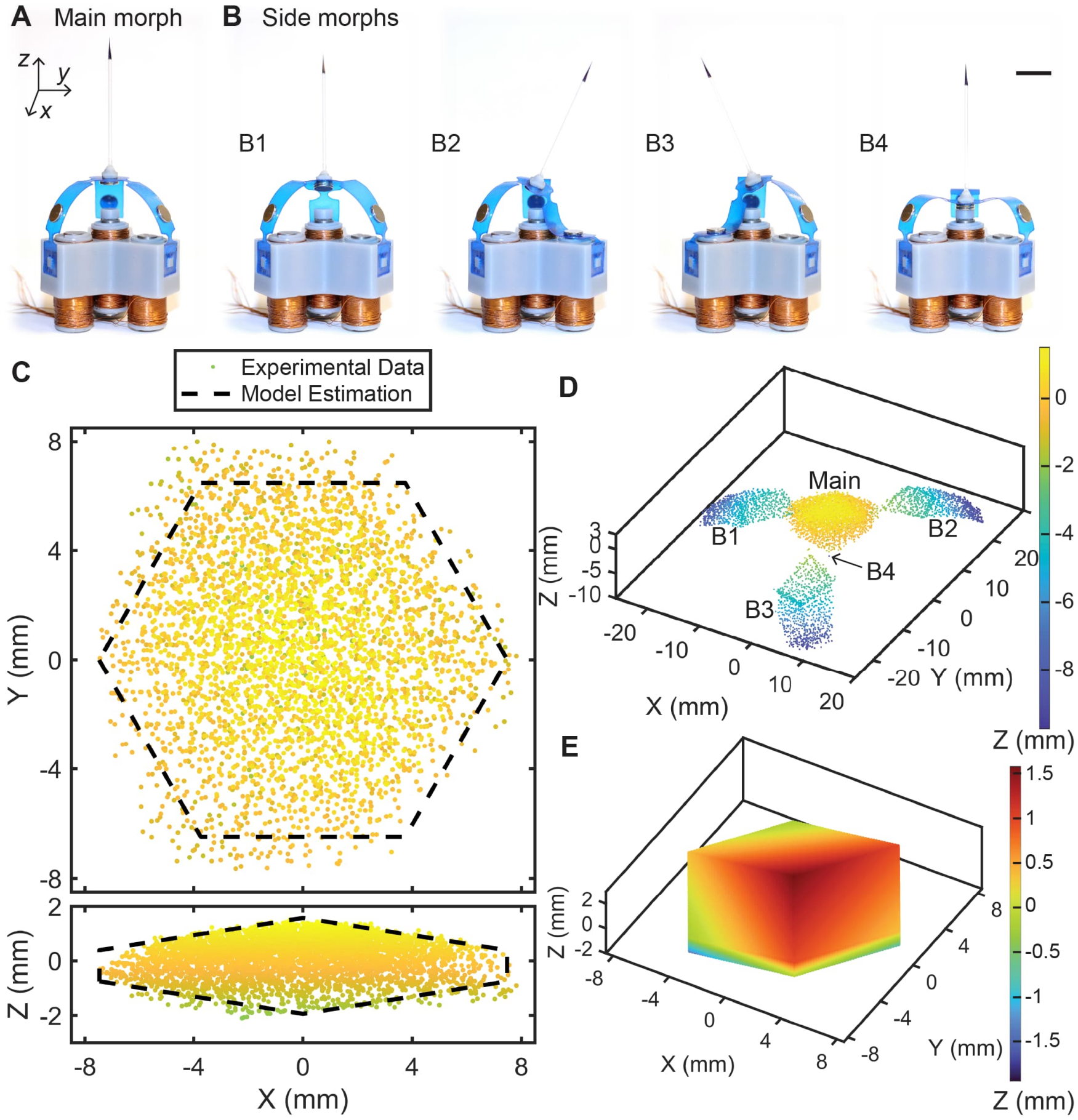

**Fig. 6.** Main morph, side morphs, respective workspaces, and blocked forces of the FilMBot. (A) Photograph of a FilMBot in its main morph, with all film legs naturally deformed and no magnet drawn in. (B) Side morphs (B1, B2, B3, and B4), each corresponds to one magnet being drawn in, scale bar 10 mm. (C) XY and XZ plane projections of the experimental and estimated workspace of the main morph. (D) 3D view workspaces of all morphs. (E) The 3D view of the estimated workspace of the main morph.

*B. Dynamic Performance*

The proposed FilMBot is distinguished by its ability to achieve high-speed movement, enabled by the low stiffness of the film-based soft kinematic structure and rapid response of electromagnetic actuation. To evaluate its maximum speed, motions of the end effector along the X, Y, Z axes, and Z-axis morph switching between side morph B4 and the main morph ($Z_{ms}$), were recorded with high-speed cameras under step signal actuation. In the Z axis, the end effector reached peak speeds of 0.54 m/s in the main morph, and 1.57 m/s during morph switching, while its rotational motion along the X and Y axes achieved maximum angular velocities up to 2117 °/s and 2456 °/s, respectively. With a 4-cm needle, the FilMBot achieved a similar linear velocity in the X and Y axes as it did along the Z axis, approximately 1.61 m/s and 1.92 m/s, respectively, both within the main morph, as illustrated in Fig. 8 and Supplementary Movie 4. The time to move from minimum to maximum position was approximately 18.5 ms (X), 15 ms (Y), 12 ms (Z), and 12 ms ($Z_{ms}$). Although these are unloaded speeds, the forces required for typical micromanipulation tasks, such as handling biological cells, microspheres, or microfibers, are generally below or near the μN level [22]–[24]. These values are negligible compared to the mN-level force capability of FilMBot.

Path-following experiments were conducted under both quasi-static and dynamic conditions using the inverse kinematic model. To examine the intrinsic precision of the micromanipulator, small-circle trajectories were executed in the central workspace region, where localized motions provide a direct measure of repeatability and the imaging system affords the resolution required for micrometer-level precision evaluation. Under the quasi-static condition, the end effector executed five cycles of circular and square trajectories in the XY plane at a frequency of one cycle per second, as illustrated in Fig. 9A. The number of repetitions was limited to five due to the memory capacity of the high-speed camera. The actual linear velocities for the circular and rectangular trajectories were approximately 2.01 mm/s and 2.56 mm/s, respectively.

The Root Mean Square (RMS) precision of the quasi-static

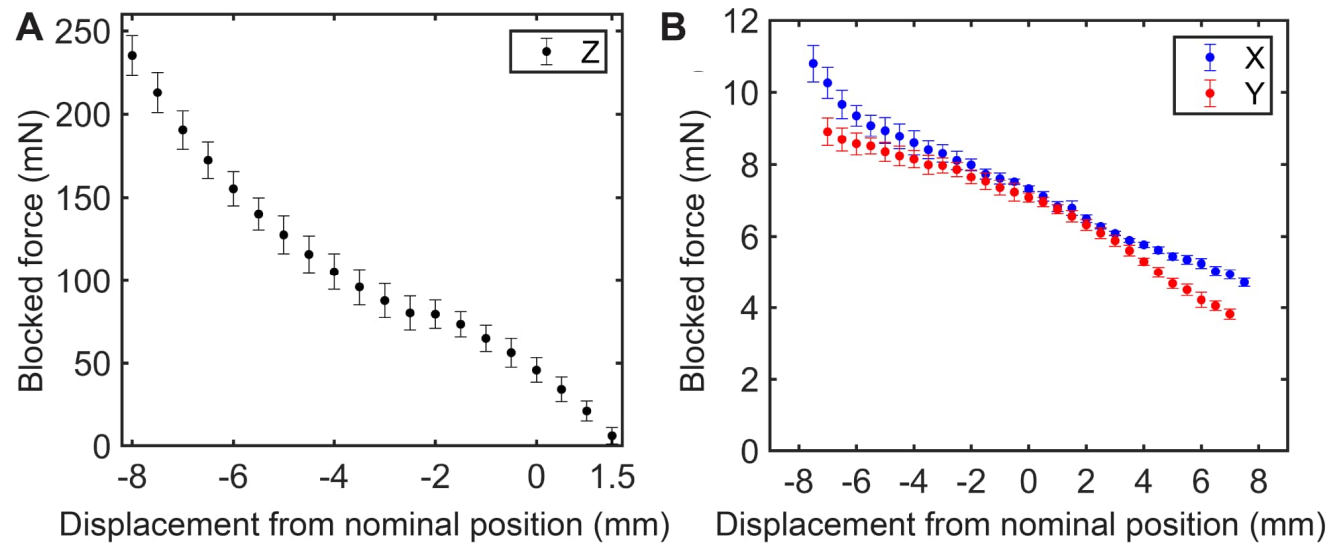

**Fig. 7.** Blocked force of the FilMBot. (A) Blocked force in the Z direction. (B) Blocked force in the X and Y directions.

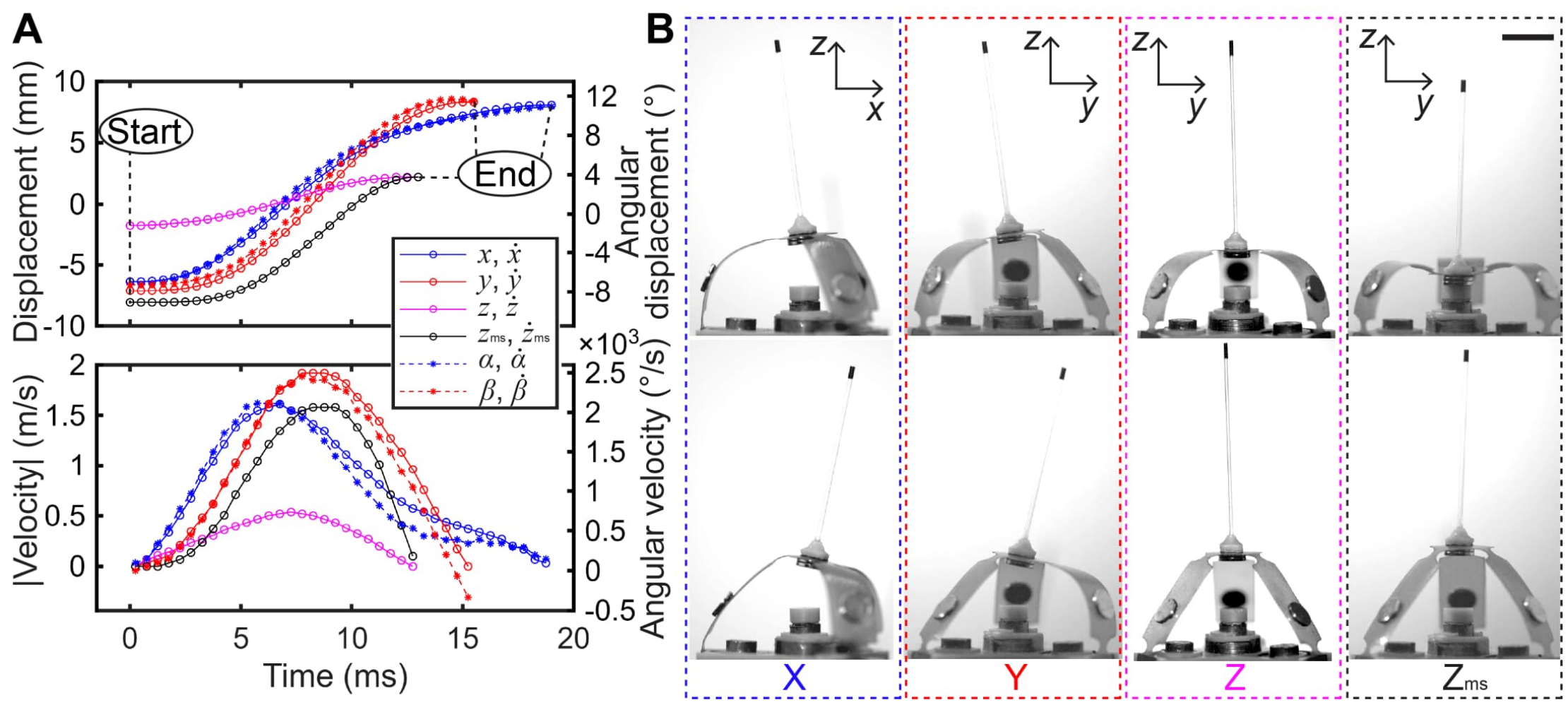

**Fig. 8.** Linear and angular displacement and respective velocity profiles of the end effector. The motions are in the main morph, with additional Z-axis morph switching between side morph B4 and main morph, $Z_{ms}$. (A) Linear displacements $x$, $y$, $z$, $z_{ms}$, and velocities $\dot{x}$, $\dot{y}$, $\dot{z}$, $\dot{z}_{ms}$, along the X, Y and Z axes; angular displacements α, β, and velocities $\dot{\alpha}$, $\dot{\beta}$, around the X and Y axes. (B) Start and end states of the FilMBot moving along the X (blue), Y (red), Z (magenta), and $Z_{ms}$ (black), scale bar 10 mm.

circular and square path-following experiments was approximately 6.3 µm and 5.7 µm, with corresponding RMS accuracy of 18.5 µm and 25.0 µm, respectively (see Table I), measured using cameras with ~6 µm/pixel resolution. These were calculated as the root mean square deviation across repeated paths (precision) and the root mean square error from the reference path (accuracy) [25], [26]. Subpixel localization [27] was employed to improve measurement precision by roughly tenfold. Considering the minimum length of the main-morph workspace in the plane as 12.96 mm, the normalized proportion of accuracy and precision regarding the workspace is approximately 0.19 % and 0.05 %, respectively.

To evaluate the dynamic performance, we conducted 0.64 mm-diameter circular path-following experiments spanning frequencies from 5 Hz to 70 Hz (see Fig. 9B), and the corresponding speed, precision, and accuracy are presented in Table I. At 5 Hz and 10 Hz, the precision and accuracy of the system remain comparable to those in quasi-static motion, although its maximum velocity increased to 20.10 mm/s at 10 Hz. As the frequency increases, the radius of the path rises slightly at 20 Hz, and reaches its peak at 30 Hz, indicating that the output energy of the micromanipulator exceeds the energy required to follow the reference. Thus, reducing the micromanipulator input currents may enhance the path-following accuracy. Notably, the average speed of the end effector is approximately 113.88 mm/s during 30 Hz operation. Upon reaching 40 Hz, the radius of the path decreases notably below the radius of the reference. This trend reverses at 50 Hz, yet the radius remains smaller than the reference. However, as the frequency increases to 60 Hz and 70 Hz, there is a significant reduction in the radius of the trajectories. Fig. 10 depicts the amplitude response observed in path-following experiments for frequencies ranging from 1 Hz to 70 Hz, using the average radius of the trajectories as the amplitude. The amplitude response peaks at 30 Hz. However, it has a valley near 40 Hz, subsequently returning to a level above -3 dB around 50 Hz, before exhibiting a gradual decline.

To evaluate the high-speed path following of FilMBot, experiments were conducted using a 12 mm diameter reference circle, as shown in Fig. 9C. The average speed reached ~1.50 m/s, with a precision of about 50 µm. We attribute the reduction in precision to the reduced spatial resolution of the imaging system, which is approximately 64 μm per pixel when imaging the outer limits of the workspace, compared to around 6 μm per pixel for the small circle and square path following experiments. Supplementary Movie 5 shows the circular motion of the FilMBot at multiple frequencies.

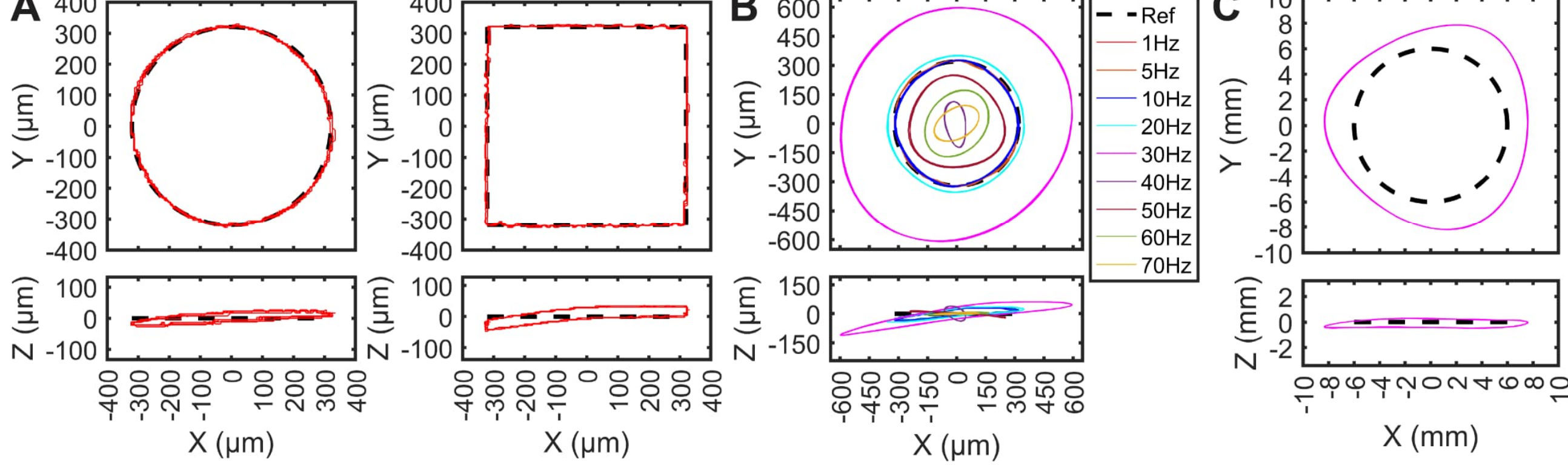

**Fig. 9.** Path following in different trajectories and frequencies. (A) Quasi-static (1 Hz) circular and rectangular trajectories, where black dashed lines are reference paths and red lines are experimental trajectories (5 cycles). Control inputs are determined using the inverse kinematic model. (B) Circular path at frequencies ranging from 5 Hz to 70 Hz, with the dashed reference path identical to the quasi-static circle reference, and solid lines representing actual trajectories (5 cycles) at various frequencies. (C) Large circular path with a mean diameter of ~16 mm at 30 Hz (5 cycles).

TABLE I
SPEED, RMS PRECISION, AND RMS ACCURACY IN PATH-FOLLOWING EXPERIMENTS.

| Path | Speed (mm/s) | RMS precision (μm) | RMS accuracy (μm) |
|---|---|---|---|
| 1 Hz circle | 2.01 | 6.3 ± 2.5 | 18.5 ± 0.3 |
| 1 Hz square | 2.56 | 5.7 ± 0.9 | 25.0 ± 0.2 |
| 5 Hz circle | 10.05 | 6.4 ± 3.2 | 24.2 ± 0.4 |
| 10 Hz circle | 20.10 | 6.0 ± 0.2 | 25.0 ± 0.4 |
| 20 Hz circle | 44.14 | 6.3 ± 3.1 | 40.9 ± 0.4 |
| 30 Hz circle | 113.88 | 6.9 ± 3.0 | 287.3 ± 1.7 |
| 40 Hz circle | 23.29 | 3.0 ± 0.8 | 238.7 ± 0.5 |
| 50 Hz circle | 76.36 | 7.1 ± 1.4 | 82.1 ± 1.8 |
| 60 Hz circle | 63.17 | 6.0 ± 1.9 | 152.5 ± 0.8 |
| 70 Hz circle | 45.35 | 7.2 ± 2.7 | 219.7 ± 0.8 |
| 30Hz large circle | 1505.07 | 50 ± 15 | 2071 ± 14 |

Furthermore, we also investigated the dynamic performance of the FilMBot by analyzing the magnitude response characteristics of individual magnet coil pairs, as shown in Appendix D. The results reveal that the magnet coil pairs, along with their associated soft kinematic structures, exhibit two resonance frequencies, approximately at 30 Hz and 50 Hz, and a magnitude response valley around 40 Hz, which exactly matches the dynamic behavior observed in the path-following experiments.

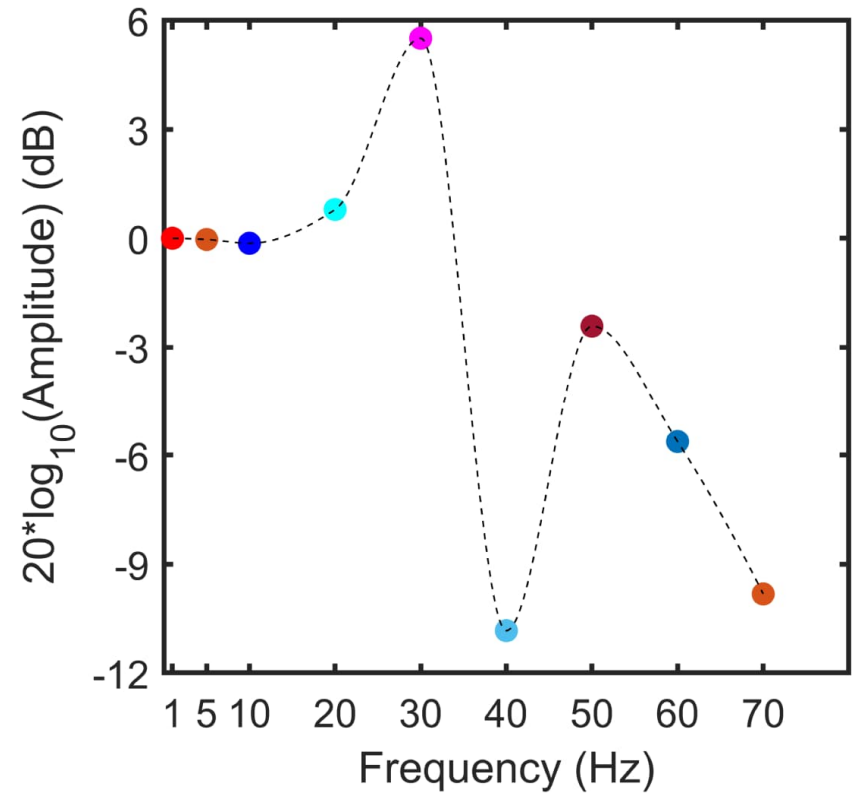

**Fig. 10.** Magnitude bode diagram of circular trajectories at different frequencies.

FilMBot is also capable of long-term operation; in our laboratory tests, the system remained functional with comparable speed and range after more than 300,000 continuous actuation cycles.

## IV. DEMONSTRATION

To demonstrate the capabilities of the FilMBot, a puncture experiment was conducted on a 15 wt. % starch gel with Young's modulus of approximately 15 kPa, as shown in Fig. 11A. The end-effector penetrated the starch gel from the bottom and reached the highest point in about 12 ms, consistent with the Z-axis displacement duration and distance traveled shown in Fig. 8. These results indicate that the FilMBot achieves a high speed along with a strong transient force, which can easily pierce the starch gel with a puncture depth of around 8.8 mm. By striking a force sensor, the peak transient force was measured to be approximately 938 mN. This level of transient force, commonly characterized as puncture force [28], is rarely reported in soft micromanipulators. It suggests the potential applications of FilMBot in biomedical sampling or therapeutic contexts requiring rapid, localized force delivery.

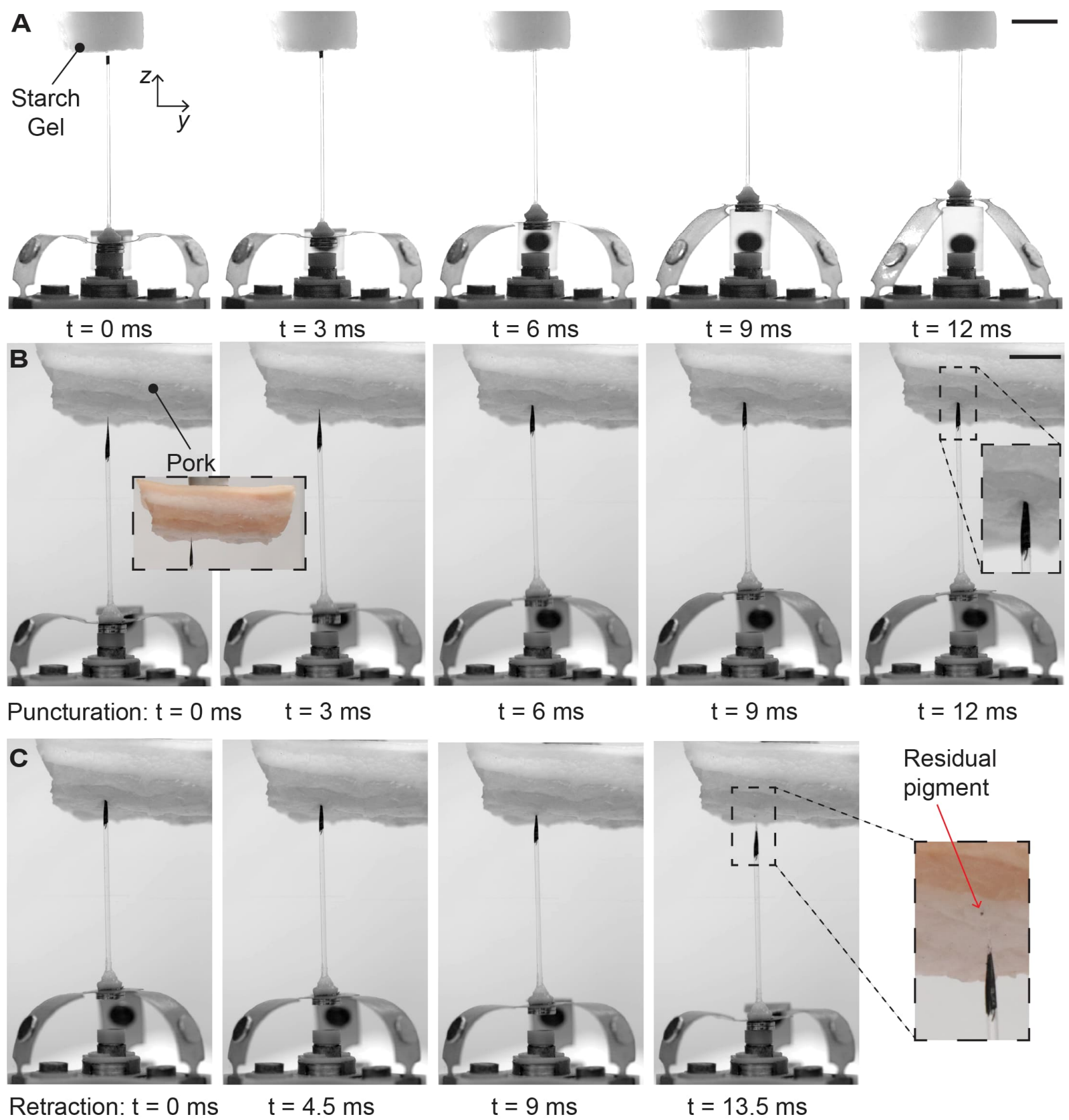

**Fig. 11.** Puncturing with the FilMBot. (A) Puncturing a 15 wt. % corn starch gel with the FilMBot. The scale bar is 10 mm. (B) Puncturation of a piece of pork, scale bar 10 mm. (C) Retraction from the pork, notice the residual pigment.

To validate its puncture ability for biological tissue, we used the FilMBot to puncture a piece of pork, as depicted in Fig. 11B. The needle tip penetrated approximately 1.0 mm into the fat layer and 2.6 mm into the muscle layer within 12 ms. The needle was then retracted, leaving the black pigment deposited in the tissue, demonstrating the potential for localized delivery, analogous to microdosing.

While most micromanipulation tasks, such as handling cells, microspheres, or fibers, require only μN-level forces, the ability of FilMBot to deliver N-level transient forces expands its applicability to more demanding biomedical procedures involving tissue interaction. This capability is particularly relevant to minimally invasive biomedical procedures such as tissue fluid extraction [29], localized diagnostics [30], and microdosing [31], where speed, precision, and force must be delivered within tight spatial and temporal constraints.

The puncture experiments for both gel and pork are shown in Supplementary Movie 6.

## V. Conclusion and Discussion

This paper presented the FilMBot, a unique film-based soft parallel robotic micromanipulator driven by noncontact electromagnetic actuation. By combining the low stiffness of the film-based soft kinematic structure with the rapid and strong noncontact electromagnetic force, FilMBot achieved an impressive speed of up to 2117 °/s, 2456 °/s and 0.54 m/s in α, β, and Z axis, respectively, and 1.57 m/s in Z-axis morph switching. Using a 4-cm needle end-effector, the robot can reach linear velocities of 1.61 m/s and 1.92 m/s along the X and Y axes, and an average velocity of ~1.50 m/s in path-following tasks. A great precision of 6.3 μm, which corresponds to 0.05 % of its workspace size, was achieved for the small path following. It has an operational bandwidth below ~30 Hz and remains responsive at 50 Hz, with performance varying with operating frequency. Additionally, FilMBot demonstrated a high transient force of around 0.93 N.

FilMBot is an α, β, Z robotic micromanipulator, where the length of the end-effector determines its performance in workspace, blocked force, linear velocity, and precision in horizontal directions. While a longer end-effector increases the horizontal linear velocity, it reduces the horizontal precision and blocked force. We argue that the 4-cm-long needle end-effector is a reasonable choice since it is similar to the size of the kinematic structure of the manipulator, gives a similar maximum linear velocity in both horizontal directions and vertical directions, a great precision, a useful blocked force, and a reasonable workspace.

With this balanced design choice, FilMBot is two orders of magnitude faster than other soft micromanipulators in the literature and four times faster than current rigid parallel micromanipulators of similar sizes, while maintaining high precision. Speeds reported in the literature for soft micromanipulators typically range from 0.001 m/s to 0.04 m/s [11]–[13]. In comparison, rigid parallel micromanipulators with similar dimensions to the FilMBot achieve higher speeds, with the examples milliDelta [16] and MiGriBot [17] reaching speeds of about 0.45 m/s and 0.067 m/s, respectively, which are particularly focused on high-speed operations. Therefore, FilMBot has a maximum speed that is over 48 times higher than that of existing soft micromanipulators and over four times faster than rigid ones reported so far. In addition, considering the possible impact of robot dimensions on speed, the speed/size ratios of the above micromanipulators are exhibited in Fig. 12, where each robot was normalized by its own largest dimension. For FilMBot, this was 9 cm, including the needle. Among them, only milliDelta [16] achieves a speed/size ratio comparable to FilMBot, though its speed is much lower at 0.45 m/s compared to 1.92 m/s from FilMBot. Soft micromanipulators generally exhibit normalized precision values ranging from 0.2% to 4% of their workspace size [11]–[13]. FilMBot, however, offers a normalized precision of 0.05 %, which is four times better, and even surpasses rigid counterparts including milliDelta [16] and MiGriBot [17] who have normalized precisions of approximately 0.16 % and 0.1 %, respectively, as shown in Fig. 12. Moreover, in addition to micromanipulation, FilMBot can exert nearly Newton-level puncture force, which was demonstrated to be sufficient to puncture biological tissues, such as fat and muscle layers [32], porcine heart [33] and bladder [34].

In addition, the FilMBot exhibits an interesting multimorph property enabled by its actuation redundancy. Different from the multimodal locomotion in mobile robots [7], [35], [36], the multimorph feature of FilMBot is primarily reflected in its adaptable operating position and extended workspaces, and all morphologies are stable and self-locking, requiring no additional system inputs to maintain them, as demonstrated in Supplementary Movie 3. It also differs from robots that use extra mechanisms to achieve multi-functionality [37], [38], FilMBot achieved multiple morphs mainly by simply utilizing the high deformability of the soft kinematic structure. This multimorph property not only broadens the range of micromanipulation tasks but also enhances the capabilities of FilMBot in rapid actions such as puncturing. The design paradigm, which adds neither size, mass, nor complexity, may inspire future soft robot designs.

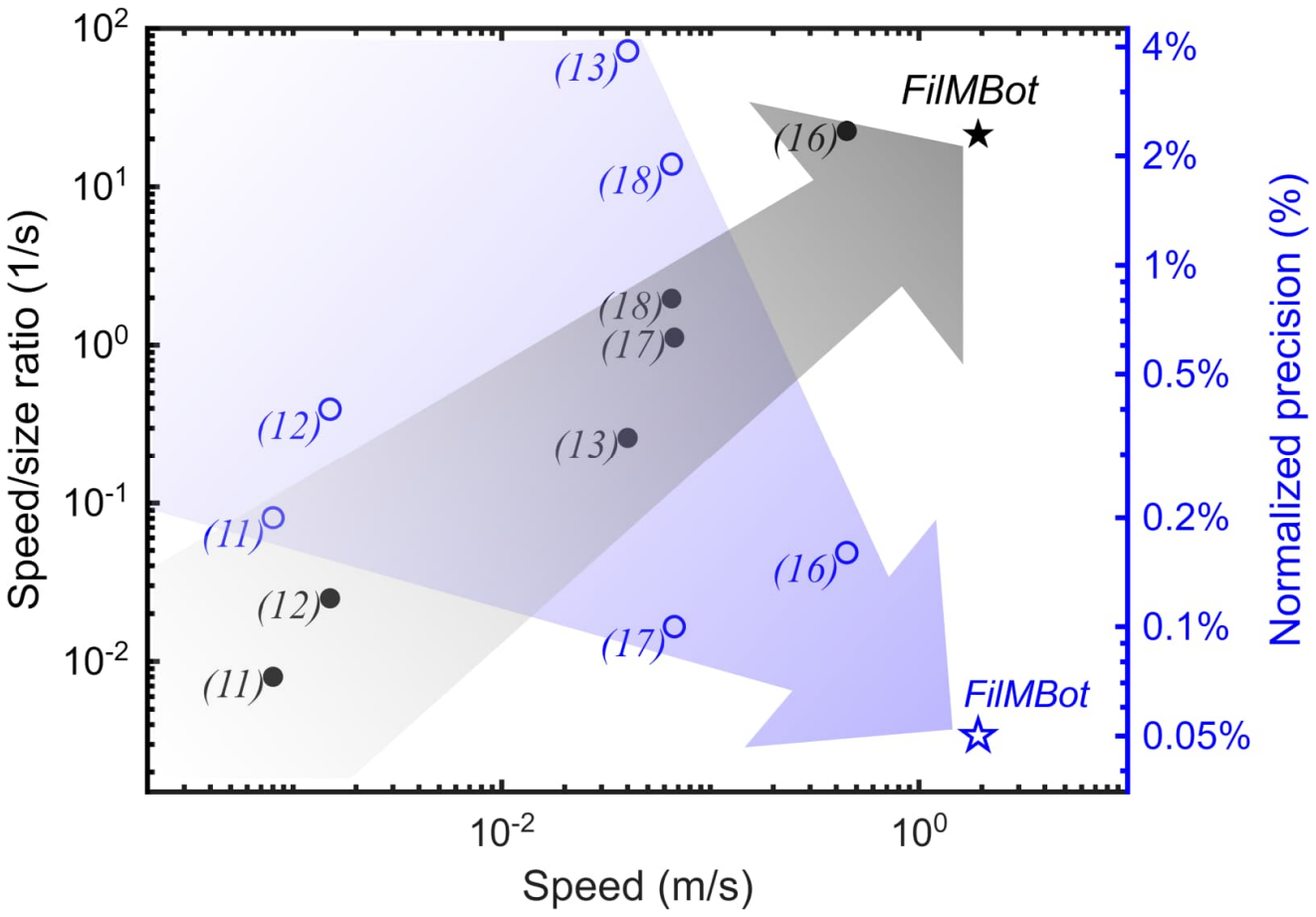

**Fig. 12.** Comparison of FilMBot with the state-of-the-art, in terms of speed, speed/size ratio, and normalized precision.

Soft robots are characterized by compliance achieved through flexible and deformable materials or mechanical properties of structures [39]–[41]. For better dynamic performances, the responsiveness of both the kinematic structure and actuator is critical. In FilMBot, the combination of low-stiffness soft kinematic structure and rapid and strong electromagnetic actuation is key in achieving its unique high-speed performance. Generally, certain actuation methods in soft robots, such as fluidic actuation [11], [13] and SMA-based actuation [14], [15], naturally respond slowly, while the combination of electromagnetic coils and permanent magnets can achieve faster responses and produce strong magnetic forces [42], [43]. However, traditional implementations using them with rigid links and joints usually encounter significant friction and mechanical resistance, which hinders speed [43], [44]. In addition, although electromagnetic actuation has been explored in soft robotics, particularly in continuum manipulators for endoscopes and catheters [45], [46], these systems typically employ stiffer materials to transmit force and torque inside organisms or complex chambers [45], resulting in lower speeds, with a reported maximum tip speed of about 5 mm/s [47]. In contrast, the FilMBot utilizes low-stiffness film-based soft kinematics that can support the weight of the structure under static conditions, while being rapidly deformed under magnetic actuation during dynamic conditions. This configuration reduces the resisting elastic force when the kinematic structure is deformed, with the maximum elastic force being approximately 16 % of the maximum magnetic force along the Z-axis, as shown in Fig. 4C. Consequently, the combined effect of the magnetic and elastic forces, along with the lightweight nature of the film, enables significant acceleration of the kinematic structure, allowing the FilMBot to achieve high speed. The workspace is also sufficiently large, providing ample distance for acceleration. Furthermore, unlike other systems that use external Helmholtz coils [7], Halbach arrays [43], or mobile coil arrays [47] for excitation, the FilMBot integrates the coils into its structure, which not only minimizes the device size but also increases the proximity between the magnets and the coil core, thereby allowing for a stronger magnetic force, contributing to the high speed of the FilMBot. Such combination and associated benefits may inspire other researchers to design future faster and more versatile soft robots.

Additionally, the FilMBot was easy to fabricate using inexpensive, readily accessible components and materials, lowering the barrier for users to construct or acquire manipulators with µm-level precision. This accessibility could broaden the application of micromanipulators beyond current academic and professional users.

FilMBot demonstrates high-speed, high-precision performance in a compact and accessible form, building on decades of insight into micromanipulation. By leveraging actuation redundancy, film-based soft kinematics, and embedded electromagnetic coils, it challenges prevailing assumptions about the speed and capability limits of soft micromanipulators. We believe this architecture offers a scalable and versatile foundation for future soft robotic systems. Looking ahead, its capabilities can be further extended through model-based optimization, accurate system modeling, and advanced control strategies, such as sensor feedback and adaptive control, broadening its applicability across diverse micromanipulation tasks.

## Appendix

### *A. Remanent magnetism and magnetic locking between coil core and magnets*

The solenoid coil core is made of S355J2 steel, which has low carbon content (≤0.20 %) and good corrosion resistance. After magnetization, the core retains certain remanent magnetism, as shown in Fig. 13, which was measured with a hall sensor (Honeywell, SS495X). Magnetic hysteresis may affect the open-loop behavior of FilMBot.

When permanent magnets are adhering to the coil core, they may be magnetically locked in place. This occurs because, at close proximity, the permanent magnet can magnetize the core exceeding the magnetization capability from the coil, resulting in the magnet not being able to be pushed away even if the coil current is reversed. In the case of two stacked magnets adhering to the core, the resulting magnetic force is greater than the elastic force generated by the deformation of the film (see Fig. 4C), leading to the locking effect. Therefore, a coil cap is needed to prevent two stacked magnets from getting too close to the central core. On the other hand, when a single magnet adhered to the core of a coil, the permanent magnet cannot magnetize the core sufficiently, so the magnet can be released from the core when the coil current is reversed since the total magnetic force is smaller than the elastic force from the deformed film.

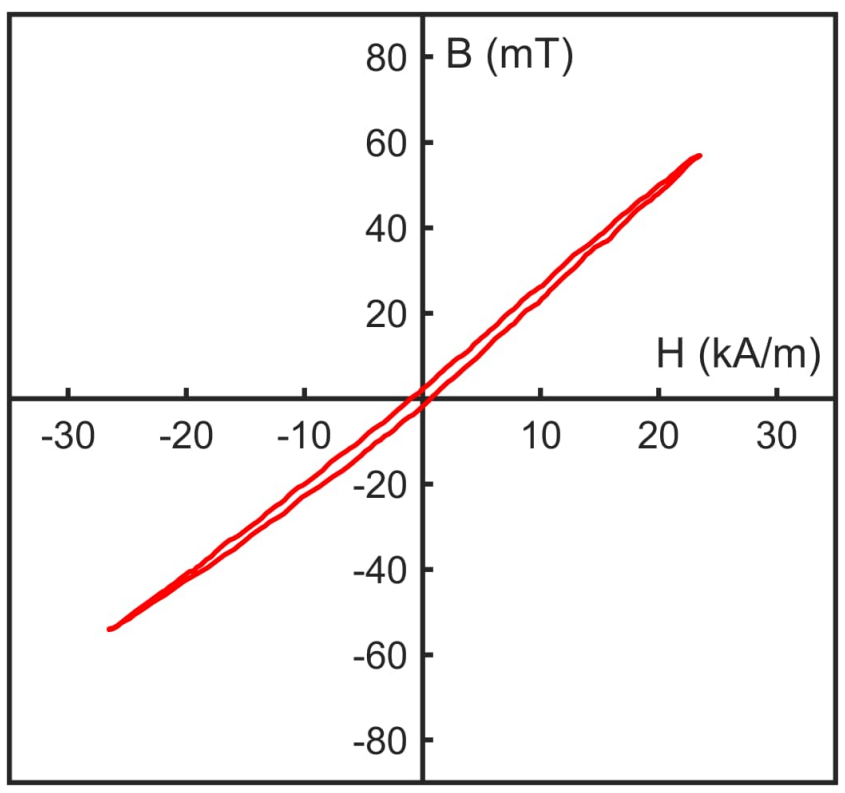

**Fig. 13.** Magnetic Hysteresis Curve of the solenoid coil with S355J2 steel core. B is magnetic flux density, and H is magnetic field strength.

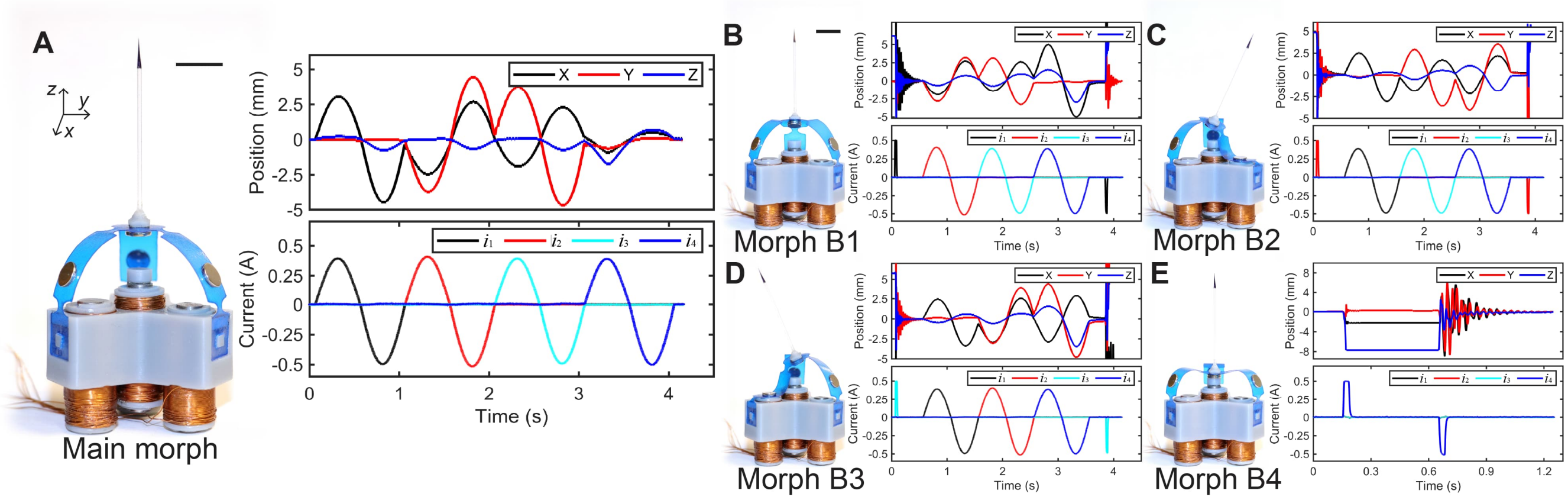

**Fig. 14.** Different morphs of the FilMBot and its response to sinusoidal currents. (A) In the main morph, the position of the end effector varies in 3D space when sinusoidal currents were applied to different coils. (B) (C) (D) The micromanipulator switches to the side morphs under a positive 0.5 A current lasting about 0.05s. Sinusoidal excitation was applied to the remaining unattached magnet coil pairs. Then, a negative 0.5 A current will make it switch back to the main morph rapidly. (E) When the top magnet was drawn in by the central coil, the workspace of the micromanipulator was limited to a single point.

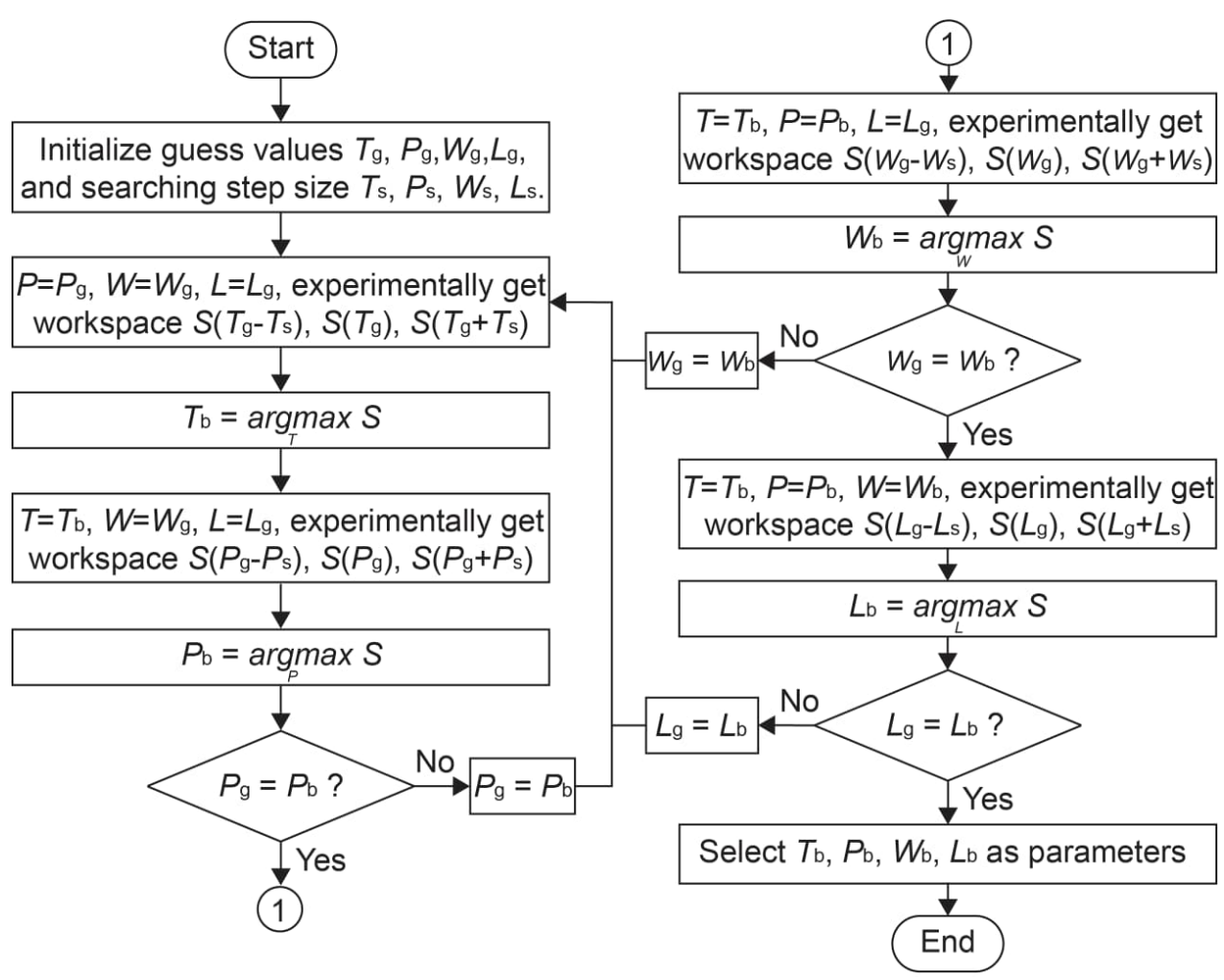

**Fig. 15.** Flowchart for iterative selection of design parameters, with the best values $T_b$, $P_b$, $W_b$, and $L_b$ yielding the largest workspace selected as the final design.

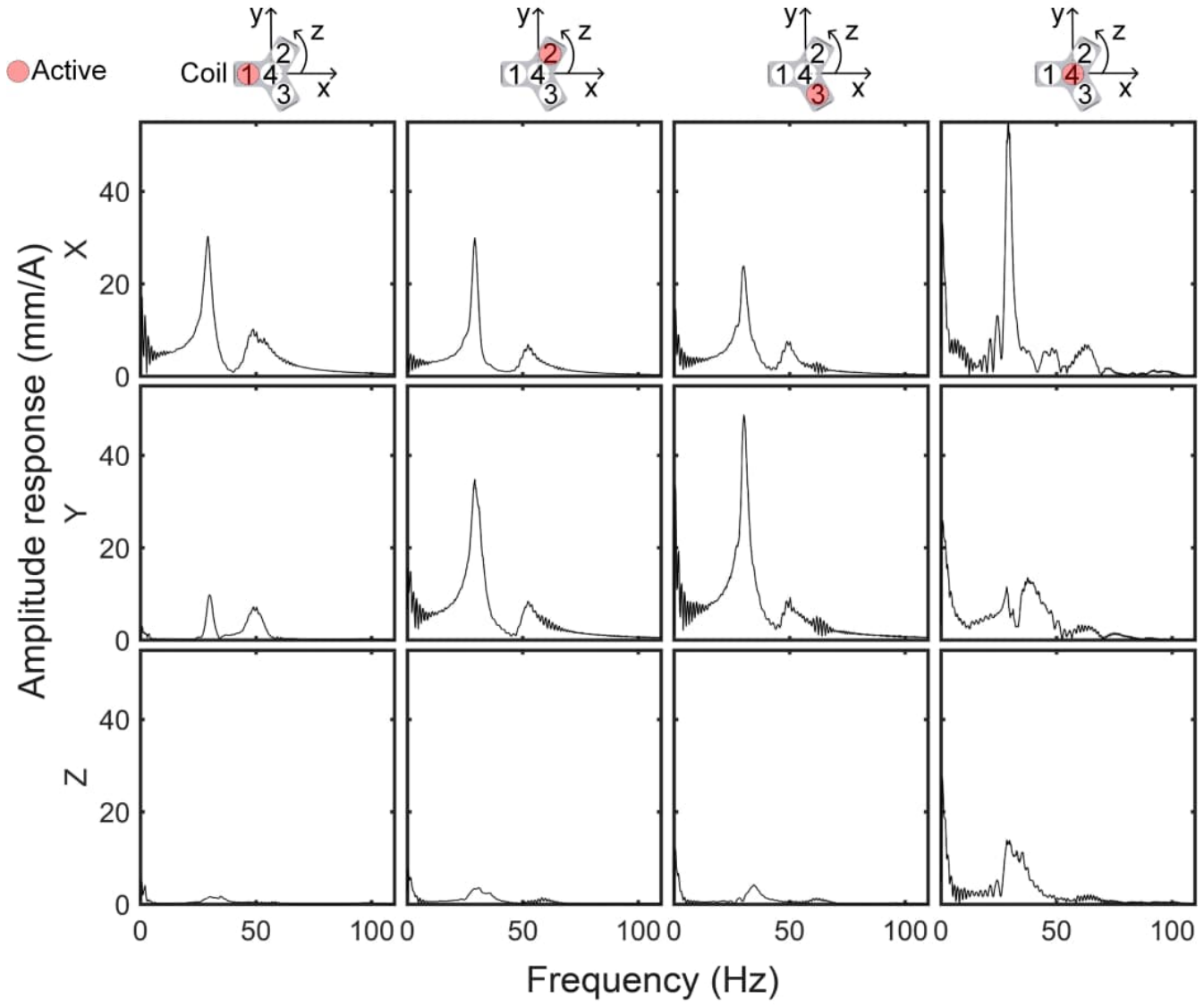

**Fig. 16.** Amplitude response of magnet coil pairs on each axis.

## *B. Motion Curves and Morph Switching*

Fig. 14 shows the response of FilMBot to sinusoidal currents in different morphs and the current signals used for morph switching, also demonstrated in Supplementary Movie 1 and 3.

## *C. Flowchart of design parameters*

The flowchart for the iterative selection of design parameters is shown in Fig. 15, corresponding to Algorithm 1.

## *D. Magnitude Response of Magnet Coil Pairs*

Chirp signals containing sinusoids ranging from 1 Hz to 100 Hz were used to excite each coil individually to characterize the system response. Position measurements were taken across the X, Y, and Z axes to identify resonant frequencies. The measurement results are detailed in Fig. 16.

## Acknowledgment

We acknowledge the use of ChatGPT (OpenAI, https://chatgpt.com) for language improvements, and all revisions were thoroughly checked by the authors to ensure the content remained unchanged.

Q.Z. led the conceptualization of the study. J.Y., H.B., H.K., and Q.Z. developed the methodology and carried out the investigation. J.Y. implemented the hardware and software, and prepared the data and visualizations. H.B. and Q.Z. acquired funding and supervised the research. J.Y., H.B., and Q.Z. performed the formal analysis and jointly wrote, reviewed, and edited the manuscript.

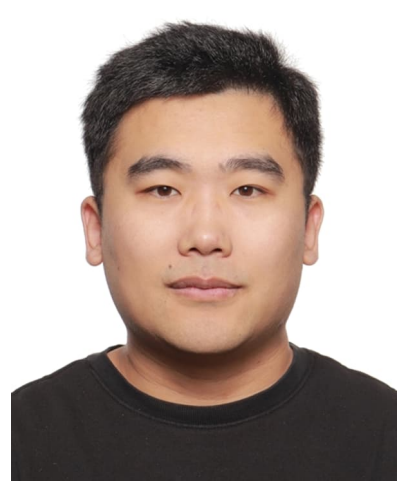

**Jiangkun Yu** received the B.Sc. degree in automation from the China University of Petroleum, Qingdao, China, in 2018, and the M.Sc. degree in instrumental science and technology from the Beihang University, Beijing, China, in 2021. He is currently working toward the Ph.D. degree with the Robotic Instruments Group, Aalto University, Finland.

His research interests include soft robotics and robotic micromanipulation.

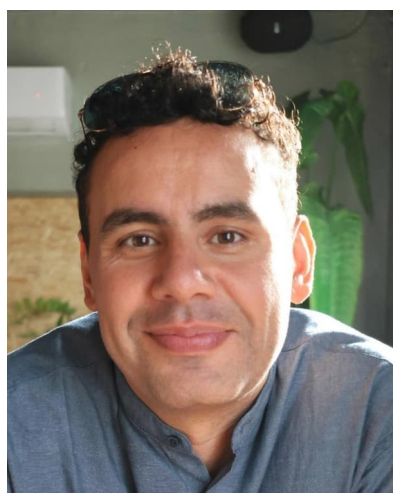

**Houari Bettahar** is an Academy Research Fellow and Principal Investigator at the Department of Electrical Engineering and Automation, Aalto University, Finland. He was a Postdoctoral Researcher (2020–2023) and a Staff Scientist (2023–2024) at Aalto University. He received his Ph.D. in Automatic Control and Robotics from FEMTO-ST Institute, University Bourgogne Franche Comté, France, in 2019, and holds a Master’s degree in Control Systems and Information Technology from the University of Grenoble Alpes, France, and a Master degree in Control Systems and Robotics from Ecole Militaire Polytechnique, Algeria.

His research interests include microrobotics and automation, micro-manipulation, mechatronics, robot control, and data-driven modeling.

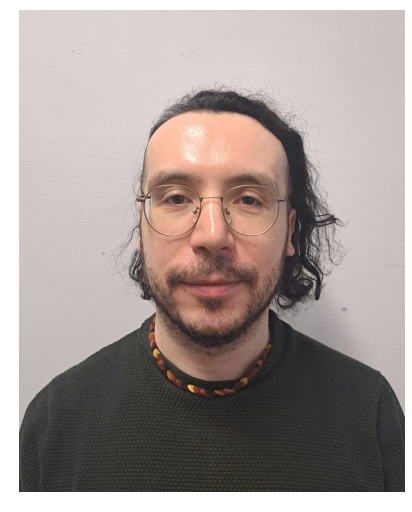

**Hakan Kandemir** received his PhD from Wageningen University in the Netherlands in 2021. His dissertation, which focused on selective particle separation in suspensions using acoustics, received the Johan Greivink PSE Award.

He is currently a Senior Scientist in the RF Microsystems group at VTT Technical Research Centre of Finland. His research interests include acoustophoresis, MEMS, and multiphysics computer modelling.

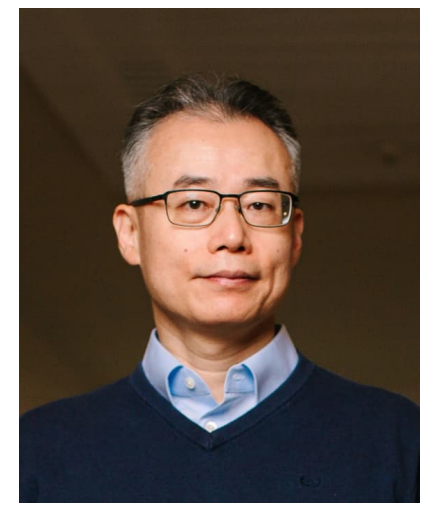

**Quan Zhou** received the M.Sc. degree in control engineering in 1996 and the Dr.Tech. degree in automation technology in 2004, both from Tampere University of Technology, Tampere, Finland.

He is currently a Professor at Aalto University, Espoo, Finland, where he leads the Robotic Instruments Group. His research interests include miniaturized robotics and robotic manipulation of small-scale objects using contact, acoustic and magnetic fields, interfacial forces, and fluidic flows.

Dr. Zhou was the recipient of the Anton Paar Research Award in 2018 and has led the EU FP7 project FAB2ASM. He has served as the Coordinator of the Topic Group on Miniaturized Robotics within euRobotics, and General Chair of the MARSS 2019 conference.